\documentclass[final]{l4dc2026}
\usepackage{enumitem}

\title[Adaptive Policy Selection and Fine-Tuning under Interaction Budgets]{Adaptive Policy Selection and Fine-Tuning under Interaction Budgets\\
for Offline-to-Online Reinforcement Learning}
\usepackage{times}
\usepackage{enumerate}
\usepackage{mathtools}
\usepackage{multirow}
\usepackage{adjustbox}
\usepackage{makecell}
\usepackage{xfrac}

\newtheorem{problem}{Problem}

\author{%
 \Name{Alper Kamil Bozkurt} \Email{bozkurta@vcu.edu}\\
 \addr Virginia Commonwealth University, Richmond, VA
 \AND
 \Name{Xiaoan Xu} \Email{xiaoan.xu@duke.edu}\\
 \addr Duke University, Durham, NC
 \AND
 \Name{Shangtong Zhang} \Email{xdm2bt@virginia.edu}\\
 \addr University of Virginia, Charlottesville, VA
 \AND
 \Name{Miroslav Pajic} \Email{miroslav.pajic@duke.edu}\\
 \addr Duke University, Durham, NC
 \AND
 \Name{Yuichi Motai} \Email{ymotai@vcu.edu}\\
 \addr Virginia Commonwealth University, Richmond, VA
}

\begin{document}

\maketitle

\begin{abstract}%
In offline-to-online reinforcement learning (O2O-RL), policies are first safely trained offline using previously collected datasets and then further fine-tuned for tasks via limited online interactions.
In a typical O2O-RL pipeline, candidate policies trained with offline RL are evaluated via either off-policy evaluation (OPE) or online evaluation (OE). The policy with the highest estimated value is then deployed and continually fine-tuned. However, this setup has two main issues. First, OPE can be unreliable, making it risky to deploy a policy based solely on those estimates, whereas OE may identify a viable policy with substantial online interaction, which could have been used for fine-tuning. Second--and more importantly--it is also often not possible to determine a priori whether a pretrained policy will improve with post-deployment fine-tuning, especially in non-stationary environments. As a result, procedures committing to a single deployed policy are impractical in many real-world settings. Moreover, a naive remedy that exhaustively fine-tunes all candidates would violate interaction budget constraints and is likewise infeasible.
In this paper, we propose a novel adaptive approach for policy selection and fine-tuning under online interaction budgets in O2O-RL. Following the standard pipeline, we first train a set of candidate policies with different offline RL algorithms and hyperparameters; we then perform OPE to obtain initial performance estimates. We next adaptively select and fine-tune the policies based on their predicted performance via an upper-confidence-bound approach thereby making efficient use of online interactions. We demonstrate that our approach improves upon O2O-RL baselines with various benchmarks.
\end{abstract}


\section{Introduction}

Dynamic physical systems that employ reinforcement learning (RL) are becoming central to complex missions across civilian, industrial, and defense domains. Examples include autonomous vehicles \citep{feng2023dense, tian2024balanced} and robotic platforms \citep{singh2022reinforcement, campanaro2024learning, chang2025spikeatac}, which increasingly operate in dynamic, nonstationary environments. By deriving decisions and control directly from onboard sensing and perception, these systems can substantially reduce operator workload and, in turn, lower the incidence of human error \citep{zhang2022reinforcement}. Deep RL \citep{tang2025deep} has been widely adopted to synthesize control policies for high-dimensional, nonlinear physical systems where manual controller design is infeasible. Despite the flexibility and power of this framework, standard RL methods typically require extensive direct interaction with the physical environment for exploration \citep{ladosz2022exploration} and are therefore impractical for training policies from scratch when such interactions are costly, risky, or time-consuming \citep{dulac2021challenges}.

Offline RL \citep{levine2020offline, prudencio2023survey} has emerged as an alternative to online RL, enabling the training of policies from large, existing datasets. These datasets are usually collected under safe, controlled conditions \citep{zhou2023real}; often, though not exclusively, by human operators. A central challenge in offline RL is that the performance of a pretrained policy becomes unpredictable as its behavior diverges from that of the policy that collected the dataset \citep{kostrikov2021offline}. Due to this distributional shift, a policy pretrained via offline RL can perform arbitrarily poorly in the real environment \citep{qin2022neorl}. To partially mitigate this issue, offline policy selection, usually via off-policy evaluation (OPE) \citep{paine2020hyperparameter, uehara2025review}, is used to identify high-performing policies among candidates pretrained under different hyperparameters. However, OPE estimates are mostly not reliable enough to determine which policy to deploy, primarily since they are also vulnerable to distributional shift \citep{brandfonbrener2021offline}, a difficulty further exacerbated by nonstationary real-world environments that demand online adaptation \citep{julian2021never}.

The necessity of online evaluation and fine-tuning of policies pretrained via offline RL has motivated the offline-to-online RL (O2O-RL) paradigm, which treats the entire pipeline as a single joint problem. By combining offline pretraining with a small amount of online interaction, O2O-RL can efficiently yield high-performing policies, enabling broader deployment of RL in physical domains. As a result, O2O-RL has attracted substantial attention and has led to the development of numerous methods. Most O2O-RL approaches \citep[e.g.,][]{nair2020awac, lee2022offline, ball2023efficient} focus on improving the offline stage to produce pretrained policies that adapt more effectively to real environments via fine-tuning. While these methods improve performance overall, they do not address the fundamental sensitivity of policy performance to hyperparameters and environments. Recently, several methods have examined policy selection in O2O-RL \citep[e.g.,][]{konyushova2021active, kurenkov2022showing}, aiming to identify the best candidate by evaluating pretrained policies under a limited budget of online interactions. Yet, these methods do not incorporate fine-tuning into policy selection and overlook the trade-offs that fine-tuning imposes on the same interaction budget.

\begin{figure}
    \centering
    \includegraphics[width=0.9\linewidth]{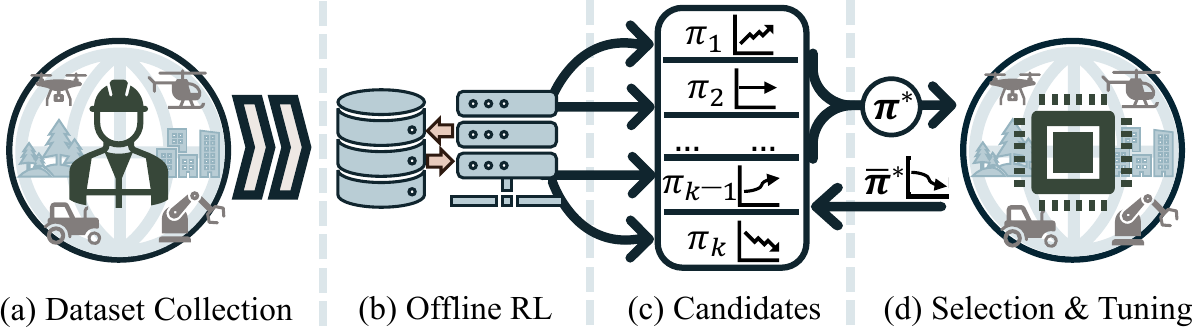}
    \caption{\textbf{Proposed O2O-RL Framework.} (a) The datasets are typically collected in controlled, task-agnostic settings. (b) Offline RL trains a diverse set of candidate policies across algorithms and hyperparameters. (c) A linear model predicts the future performance of each policy with a UCB. (d) The policy with the highest UCB is selected and fine-tuned; its predicted performance and UCB are updated during fine-tuning; whenever its UCB falls below that of another candidate, the selected policy is replaced by that candidate.}
    \label{fig:framework}
\end{figure}

In this work, we address the problem of obtaining high-performing, deployable policies in the context of O2O-RL from a broader and more practical perspective. Prior work and our experiments show that pretrained policies can perform arbitrarily poorly in real environments, and that fine-tuning may require substantial interaction to improve performance and can even cause regressions, with outcomes varying across algorithms, hyperparameters, and environments. To manage this volatility in performance under interaction budgets, we propose, to our knowledge, the first approach (Figure~\ref{fig:framework}) that jointly combines adaptive policy selection with fine-tuning. In the offline stage of the approach, we first train a diverse set of candidate policies with offline RL, spanning a representative range of algorithms and hyperparameters, and then use OPE to obtain initial performance estimates, similar to \cite{konyushova2021active}. In the novel online stage, we actively select and fine-tune candidate policies using an upper-confidence-bound (UCB) criterion on their predicted future performance until the interaction budget is exhausted. After each online episode, we evaluate the performance of the policy, then update the prediction and the confidence bound for the selected policy via a linear autoregressive (AR) model. We adaptively switch policies when the bound for the selected policy falls below that of another candidate policy. We evaluate our approach on multiple benchmarks and show consistent improvements over O2O-RL baselines.

\section{Related Work}

\subsection{Offline Reinforcement Learning} 

\paragraph{Basic Approaches.}
A naive approach to offline RL is to train policies via standard online, off-policy algorithms such as deep deterministic policy gradient \citep[DDPG,][]{lillicrap2016continuous}, twin delayed DDPG \citep[TD3,][]{fujimoto2018addressing}, and soft actor-critic \citep[SAC,][]{haarnoja2018soft} on batches drawn from previously collected datasets. However, such policies can fail catastrophically, as they may lead to states outside the dataset due to substantially inaccurate value estimates arising from distributional shift.
In contrast, behavior cloning \citep[BC; e.g.,][]{torabi2018behavioral, florence2022implicit}, which aims to imitate the dataset behavior and thereby largely prevent distributional shift, often cannot surpass the performance of the behavior policy, which is undesirable for non-expert datasets.

\paragraph{Offline RL for Distributional Shift.}
The central objective in offline RL is to improve upon the behavior policy of the dataset while avoiding catastrophic failures caused by distributional shift. Most offline RL methods therefore balance this trade-off by permitting off-policy updates with additional regularization or constraints. Some approaches, such as batch-constrained deep Q-learning \citep[BCQ,][]{fujimoto2019off}, bootstrapping error accumulation reduction \citep[BEAR,][]{kumar2019stabilizing}, and policy in the latent action space \citep[PLAS,][]{zhou2021plas}, focus on constraining actions to prevent policies from taking actions outside the dataset. Other approaches, such as conservative Q-learning \citep[CQL,][]{kumar2020conservative} and TD3+BC \citep{fujimoto2021minimalist}, incorporate, respectively, Q-value and BC regularization terms into the policy updates. We refer readers to \cite{levine2020offline} and \cite{prudencio2023survey} for comprehensive reviews of offline RL approaches.

\subsection{Offline-to-Online Reinforcement Learning}

\paragraph{Offline RL for Fine-Tuning.}
The fundamental limitation that the performance of offline RL methods is upper-bounded by dataset quality necessitates fine-tuning pretrained policies via online interactions. This insight has motivated methods that treat the entire offline-to-online framework as a unified problem, where the main objective is to improve the final performance of policies after fine-tuning. Advantage weighted actor critic \citep[AWAC,][]{nair2020awac} improves the efficiency of fine-tuning with off-policy data via dynamic programming. Implicit Q-learning \citep[IQL,][]{kostrikov2021offline} utilizes a policy-extraction step during training that aids fine-tuning. Calibrated Q-learning \citep[CalQL,][]{nakamoto2023cal}, building on CQL, learns conservative Q-values better suited for fine-tuning. Revisited behavior regularized actor-critic \citep[ReBRAC,][]{tarasov2023revisiting} extends TD3+BC with simple hyperparameter choices that benefit both offline training and fine-tuning. Hybrid RL \citep[Hy-Q,][]{song2023hybrid} augments offline and online data and applies a fitted Q-iteration procedure. In this work, we do not attempt to improve any specific offline RL algorithm. Instead, we treat the choice of algorithm as a high-level hyperparameter when training candidate policies and fine-tuning, since no single offline RL method is uniformly best in all environments.

\paragraph{Online Policy Selection.}
A recent line of work, \cite{konyushova2021active} and \cite{kurenkov2022showing}, studied the policy selection problem in O2O-RL. In a setting similar to ours, these studies use a limited online-interaction budget to identify the best-performing pretrained policies, rather than relying on OPE estimates as in prior offline RL approaches \citep{paine2020hyperparameter, uehara2025review}. Since evaluating all candidates online would quickly exhaust the budget, they propose active selection strategies to decide which policies to evaluate, thereby allocating the budget efficiently. In contrast, our focus is to allocate online interactions to identify the policies that will perform best after fine-tuning under the given budget.

\section{Adaptive Policy Selection and Fine-Tuning}

\subsection{Problem Formulation}
Following the standard RL framework, we model the interaction with an environment to perform tasks as Markov decision processes (MDPs). Formally, an MDP $\mathcal{M}$ is a tuple $(S,A,P,p_0,R,\gamma)$ where $S$ is the state set, $A$ is the action set, $P$ and $p_0$ are the transition and the initial-state distributions, $R$ is the reward function, and $\gamma$ is the discount factor. The ultimate objective in RL is to obtain a policy $\pi:S\mapsto A$ maximizing the expected cumulative discounted reward; i.e., $\pi^* \coloneqq \mathrm{argmax}_\pi \mathbb{E}\left[\sum_{t=0}^\infty \gamma^t r_t\right]$. In offline RL, this objective must be achieved using a dataset $\mathcal{D}$ of transitions $\{(s_i,a_i,r_i,s'_i)\}_{i=1}^M$ collected by a typically unknown behavior policy $\pi_b$ in $\mathcal{M}$ without further interaction with the environment.

In this work, we extend the offline RL setting by allowing a limited budget $N$ of online interactions in addition to the dataset $\mathcal{D}$; i.e., a set of at most $N$ new transitions may be collected from the environment. Unlike \cite{konyushova2021active} and \cite{kurenkov2022showing}, in our O2O-RL setting these online interactions can be used not only for evaluation but also for fine-tuning, introducing a novel trade-off. Given an MDP $\mathcal{M}$, a dataset $\mathcal{D}$ collected from $\mathcal{M}$, and an interaction budget $N$, our aim is to yield the highest performing policy achievable, in the sense of minimizing regret. We formally define our specific problem as follows:

\begin{problem} \label{problem}
Consider a set of $K$ candidate policies trained offline on a dataset using different hyperparameter settings, and let $\pi^i_j$ denote the policy that is obtained from the $i$th pretrained policy after $j$ fine-tuning   iterations, and let $v^i_j \coloneqq \mathbb{E}_{\pi^i_j}\left[\sum_{t=0}^\infty \gamma^t r_t\right]$ denote its corresponding value. For a given interaction budget $N$, the optimal selection is then a policy $\pi^{i^*}_{j^*}$ such that $i^*, j^* \coloneqq \underset{1<i<K,0<j\leq N}{\mathrm{argmax}} v^i_j$. Our objective is to devise a procedure that fine-tunes each pretrained policy $\pi^i_0$ for $\bar{j}_i$ iterations and estimates the values $\{\hat{v}^i_0, \dots, \hat{v}^i_{\bar{j}_i}\}$ under the constraint $\sum_{i=1}^K \bar{j}_i \leq N$ such that the value of the selected policy $\hat{v} = \max\limits_{1<i<K} \max\limits_{0<j\leq \bar{j}_i} \hat{v}^i_j$ minimizes $\textrm{regret} = v^{i^*}_{j^*} - \hat{v}$.
\end{problem}

\subsection{Main Approach}
Our overall procedure is shown in Algorithm \ref{alg}. We now explain our approach in detail below.

\paragraph{Offline Initialization.} We follow the standard O2O pipeline. First, we train a diverse pool of $K$ candidate policies $\{\pi_0^i,\dots, \pi_0^K\}$ with offline RL on the given dataset $\mathcal{D}$ by sweeping across multiple algorithmic families and hyperparameter ranges to cover the design space suitable for the environment $\mathcal{M}$. We then perform OPE to estimate the value $\hat{v}_\textrm{OPE}^i$ of each pretrained policy $\pi_0^i$ using off-the-shelf methods. As we show in our experiments, these OPE estimates could be substantially inaccurate but are often correlated with the actual values. Thus, we use these estimates to rank the policies for initial policy selection and do not directly incorporate them into our fine-tuning value estimation. Instead, we initialize the estimates $\hat{v}_0^i$ with the value estimate $\hat{v}_B$ of the behavior policy $\pi_B$ calculated using the transitions in $\mathcal{D}$.

\LinesNumbered
\SetAlCapHSkip{0pt} %
\setlength{\algomargin}{1em}

\begin{algorithm2e}[h]
\SetKwInOut{Input}{Input}\SetKwInOut{Output}{Output}
\SetKwComment{Comment}{}{}
\Input{an MDP $\mathcal{M}$, a dataset $\mathcal{D}$, an interaction budget $N$}
\Output{a policy $\pi^*$}
\caption{Adaptive Policy Selection and Fine-Tuning}\label{alg}
\small
\Comment{\vspace{-0.7em}}

\Comment{\# Offline Stage}
Estimate behavior policy value $\hat{v}_B$ using $\mathcal{D}$\\
Train $K$ candidate policies $\{\pi_0^i\}_{i=1}^K$ on $\mathcal{D}$ with diverse offline RL algorithms/hyperparameters\\
Estimate the initial policy values $\{\hat{v}_\textrm{OPE}^i\}_{i=1}^K$ via OPE\\
Argsort the policies based on $\{\hat{v}_\textrm{OPE}^i\}_{i=1}^K$ to obtain OPE ranks $\{r^i\}_{i=1}^K$\\
Initialize value list $\hat{\mathbf{v}}^i{\leftarrow}(\hat{v}^i_{-\underline{t}},\dots,\hat{v}^i_{0})$ with pseudo-observations $\hat{v}_B$ for each $\pi_0^i$\\
$\mathrm{heap} \leftarrow \mathrm{PriorityQueue}(\{(\pi_0^i, \hat{v}_B, r^i)\}_{i=1}^K)$ \Comment{{\footnotesize \# prioritizes values over OPE ranks}}
$\Pi \leftarrow \{ \}$ \Comment{{\footnotesize \# policy storage}}
\Comment{\vspace{-0.7em}}

\Comment{\# Online Stage}
\For{$j=1$ \KwTo $N$}{
\Comment{\vspace{-0.8em}}

\Comment{\small{\# Fine-Tuning \& Evaluation}}
$\pi^i_t \leftarrow \mathrm{heap.pop}()$\\
$\pi^i_{t+1} \leftarrow \mathrm{finetune}(\mathcal{M}, \pi^i_t)$\\
$\hat{v}^i_{t+1} \leftarrow \mathrm{evaluate}(\mathcal{M}, \pi^i_{t+1})$\\
$\Pi\mathrm{.append}((\hat{v}^i_{t+1},\pi^i_{t+1}))$\\
$\mathbf{v}^i\mathrm{.append}(\hat{v}^i_{t+1})$\\
\Comment{\vspace{-0.9em}}

\Comment{\small{\# Value Forecast}}
Fit an AR(2)-ARCH(1) model $m$ to $\mathbf{v}^i$ \\
$\tau \leftarrow \{t{+2, \dots, t{+}1{+}N{-}j}\}$ \Comment{{\footnotesize \# forecast iteration numbers}}
Simulate values $\{(\tilde{v}^{(s)}_{t'})_{t'\in \tau}\}_{s=1}^R$ using the fitted $m$\\
Compute the UCB (the 95th percentile) value $\bar{v}^i_{t'}$ from $\{\tilde{v}^{(s)}_{t'}\}_{s=1}^R$ for all $t'\in\tau$\\
$\bar{v}^i_* \leftarrow\max\limits_{t'\in\tau}\bar{v}^i_{t'}$\\
$\mathrm{heap.push}((\pi^i_{t+1},\bar{v}^i_*, 0))$  \Comment{{\footnotesize \# zero implies higher priority than OPE orders}}
}
$\pi^* \leftarrow \mathrm{max}\{\Pi\}$ \Comment{{\footnotesize \# max based on value estimates}}
\Return{$\pi^*$}
\end{algorithm2e}

\begin{figure}
\centering
\includegraphics[width=\linewidth]{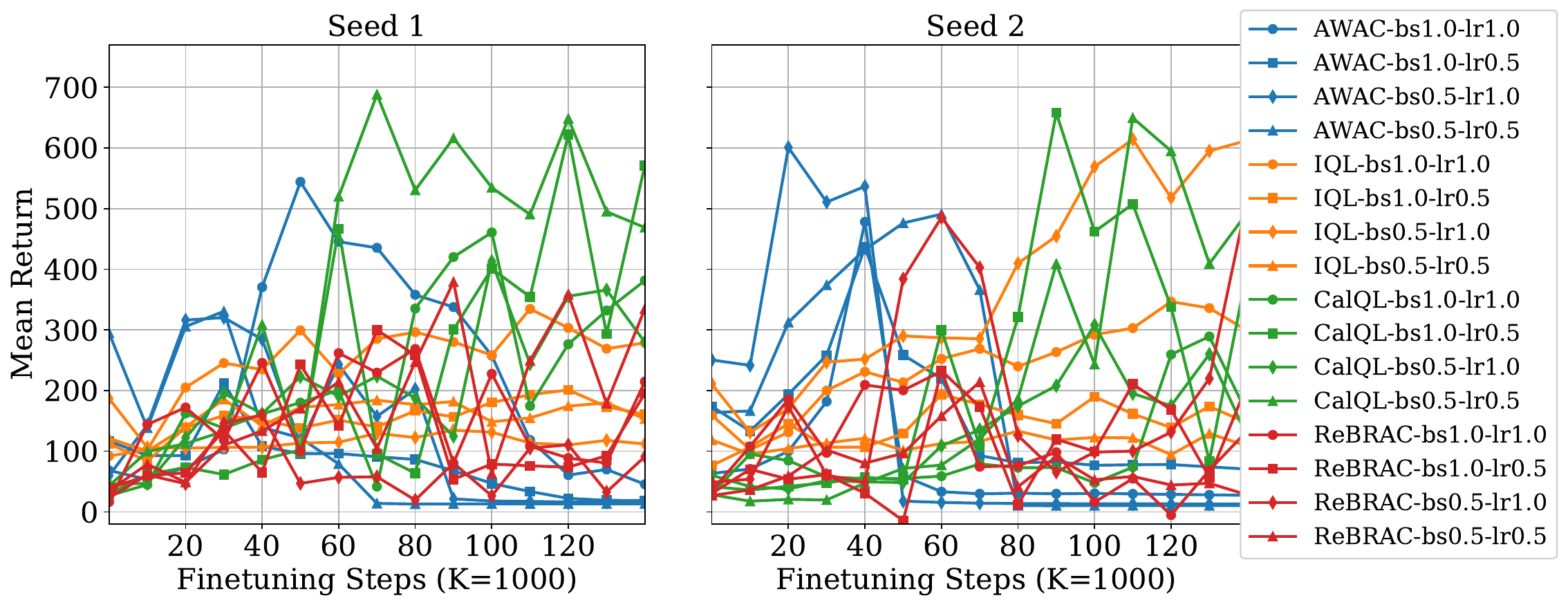}
\caption{
Evolution of the mean return values of pretrained policies during fine-tuning on \textsc{walker-random} for two random seeds. Policies are pretrained offline for 200 K steps using default and half-default batch sizes (bs) and learning rates (lr). Values are obtained by averaging returns over 100 rollouts. The value curves are highly irregular: they may improve (e.g., CalQL), regress after initial improvement (e.g., AWAC), or exhibit high variance. Even with the same algorithms and hyperparameters, changing the seed can significantly alter the outcome of offline training and the progression of the value curves, which necessitates an adaptive policy selection approach.
}
\label{fig:finetuning}
\end{figure}

\paragraph{Fine-Tuning Value Evolution Model.} As we demonstrate in Figure \ref{fig:finetuning}, the values of policies during fine-tuning do not always monotonically increase; they may stall or regress, exhibit sudden drops or jumps, and pass through both high- and low-variance regimes. We model this nonstationarity in value during fine-tuning with an AR process combined with a conditionally heteroscedastic (ARCH) \citep{engle1982autoregressive} process. In particular, we adopt a linear AR(2) model with two lags and the standard ARCH(1) model with only one lag to capture the short-term upward/downward trends and time-varying variance, respectively. While we favor this minimalist design to highlight the key aspects, other richer models such as AR with (integrated) moving-average \citep[ARMA, ARIMA,][]{box2015time}, or generalized ARCH \citep[GARCH,][]{bollerslev1986generalized} models with higher lags can also be employed, depending on the application. Formally, we consider the following model:
\begin{align}
    \mu_t &= \beta_0 + \beta_1\hat{v}_{t-1}^i + \beta_2\hat{v}_{t-2}^i \label{eq:mean}\\
    \hat{v}_t^i &=  \mu_t + \epsilon_t \\
    \epsilon_t &= \sigma_t \eta_t \\
    \sigma^2_t &= \alpha_0 + \alpha_1 \epsilon^2_{t-1} \label{eq:variance}
\end{align}
where $\eta_t \overset{\mathrm{iid}}{\sim} N(0,1)$ are identically and independently distributed standard normal residuals. Here, $\beta_0, \beta_1, \beta_2$, and $\alpha_0, \alpha_1$ are mean and variance parameters to be estimated from observed values. The standard methods such as ordinary least squares (OLS) or quasi-maximum likelihood (QML) can be used to estimate these variables using off-the-shelf tools. To mitigate the initial identification problem due to the lack of early observations, we start with $\underline{t}$ pseudo-observations $(\hat{v}^i_{-\underline{t}},\dots,\hat{v}^i_{0})$, each set to $\hat{v}_B$, with $\underline{t}$ chosen according to model complexity. We refer readers to \cite{box2015time} and \cite{francq2019garch} for methodological estimation details.

\paragraph{Fine-Tuning Value Forecast.} After each fine-tuning iteration, we re-estimate the parameters of AR(2)-ARCH(1) accounting for the new observed value estimate and then use these estimated parameters to forecast the future values. Although the mean and the variance in \eqref{eq:mean} and \eqref{eq:variance} admit closed-form multi-step forecasts, we rely on simulation to keep our framework more compatible with general models. Concretely, for a fine-tuned policy $\pi^i_t$ and a remaining budget $n$, we simulate $R$ sequences of values $\{(\tilde{v}^{(s)}_{t+1},\dots,\tilde{v}^{(s)}_{t+n})\}_{s=1}^R$ with residuals drawn from a standard normal distribution. For each future fine-tuning iteration $t' \in \{t{+}1,\ldots,t{+}n\}$, we compute the 95th percentile of $\{\tilde{v}^{(1)}_{t'}, \dots, \tilde{v}^{(R)}_{t'}\}$, and use these percentiles as UCBs, denoted $\bar{v}^i_{t'}$, on the forecasted values. We then calculate the maximum UCB (max-UCB), $\bar{v}^i_* = \max\limits_{t'=t+1,\dots,t+n}\bar{v}^i_{t'}$, and push the policy $\pi^i_t$ with its max-UCB $\bar{v}^i_*$ to a max-priority queue data structure.

\paragraph{Policy Selection for Fine-Tuning.} The online stage proceeds as an iterative fine-tuning loop. At each iteration, we pop from a priority queue the policy $\pi_t^i$ with the largest max-UCB, fine-tune it to obtain $\pi_{t+1}^i$, evaluate $\pi_{t+1}^i$ via online interaction, and store $\pi_{t+1}^i$ with its estimated value $\hat{v}_{t+1}^i$ in the policy set $\Pi$. We then forecast the values, compute the updated max-UCB $\bar{v}*^i$, and push $\pi_{t+1}^i$ back into the queue with $\bar{v}*^i$. Once the interaction budget is exhausted, we return the policy with the largest estimated value from the policy set $\Pi$. To incorporate OPE estimates into selection, we rank the OPE scores and initialize the queue by pushing them alongside $\hat{v}_B$, with the queue ordering prioritizing max-UCBs over OPE ranks. This allows OPE to guide early choices, since all policies initially share the pseudo max-UCB $\hat{v}_B$. Fine-tuned policies are pushed with the highest OPE rank to maintain consistency and to prioritize them over policies that have not yet been fine-tuned whenever their max-UCB is greater than or equal $\hat{v}_B$.

\section{Experiments}

\subsection{Implementation Details}

We evaluate our approach on four standard legged-robot locomotion tasks from the \textsc{D4RL} offline RL benchmark \citep{fu2020d4rl}, implemented in \textsc{PyBullet} \citep{coumans2021}: \textsc{hopper}, \textsc{(half)cheetah}, \textsc{walker(2d)}, and \textsc{ant}, where each non-terminated trajectory contains 1,000 transitions. For each environment, we use four datasets (16 in total), covering increasing levels of expertise, namely, \textsc{random}, \textsc{medium}, \textsc{medium-replay}, and \textsc{medium-expert}, each containing 1,000 K (K=1,000) transitions. See \citep{farama2021} for details on how these datasets are generated.

We employ four representative training algorithms designed for offline pretraining followed by online fine-tuning: AWAC \citep{nair2020awac}, IQL \citep{kostrikov2021offline}, CalQL \citep{nakamoto2023cal}, and ReBRAC \citep{tarasov2023revisiting}. We adopt the implementations of these algorithms provided by the offline RL library d3rlpy \citep{d3rlpy}. For each algorithm, we consider four hyperparameter settings from the cartesian product of two batch sizes (default and half-default) and two learning rates (default and half-default). We pretrain a candidate policy for each setting for 200 K iterations on datasets, with all remaining hyperparameters fixed to the d3rlpy defaults, yielding 16 pretrained candidate policies in total. We then utilize the Fitted Q-Evaluation \citep[FQE,][]{le2019batch} implementation from d3rlpy to obtain OPE estimates of these pretrained policies, alongside the estimates produced by their respective critic networks. We fine-tune the pretrained policies with online interactions with the environments using the same hyperparameters used to pretrain them. We use the Python library \textsc{arch} \citep{arch} to estimate the AR(2)-ARCH(1) model parameters and forecast the future values of policies. 

\subsection{Main Results}

We compare our adaptive policy selection and fine-tuning approach with several standard O2O-RL baselines.
\textbf{\texttt{OPE}}: The pretrained policy with the highest OPE estimate is selected without fine-tuning.
\textbf{\texttt{Best}}: The pretrained policy with the highest true value is selected without fine-tuning (hypothetical upper bound for reference). 
\textbf{\texttt{OE}} (Online Evaluation): All pretrained policies are evaluated online by sharing the interaction budget equally across them; the policy with the highest estimated value is then selected without fine-tuning.
\textbf{\texttt{FT}}: The policy selected by \textbf{\texttt{OPE}} is fine-tuned using the entire interaction budget.
\textbf{\texttt{Ours}}: See Algorithm~\ref{alg}. The value lists are initialized with $\underline{t}{=}5$ pseudo-observations. In each fine-tuning iteration, 10 K online transitions are used for fine-tuning and another 10 K for evaluation, for a total cost of 20 K transitions per iteration. The values are forecast and the UCBs are computed with $R=100$ simulations.

Our main results for online interaction budgets of 160 K and 320 K are presented in Tables \ref{table:160} and \ref{table:320}, respectively. We compare O2O-RL approaches by evaluating the policies they select, using the mean return over 100 rollouts. For each environment, we rescale the mean returns using min-max normalization, with the minimum set to the value estimate of a random policy on the dataset, and the maximum set to the value of the highest-performing policy ($v^{i^*}_{j^*}$; see Problem \ref{problem}) that would be obtained if future fine-tuning values were known a priori. The experiments are repeated with four seeds and scores are reported as mean values with associated standard deviations, expressed as percentages.

\begin{table}[htbp]
\begin{center}

\caption{Results for O2O-RL baselines and our approach under a 160 K online interaction budget.} \label{table:160} 
\begin{adjustbox}{width=0.81\textwidth}
\begin{tabular}{cc|cc|ccc}
\multicolumn{2}{c|}{\multirow{2}{*}{Environment}}  & \multicolumn{2}{c|}{Offline} & \multicolumn{3}{c}{Online} \\ \cline{3-7} 
 & & \multicolumn{1}{c|}{\texttt{OPE}} & \multicolumn{1}{c|}{\texttt{Best}} & \multicolumn{1}{c|}{\texttt{OE}} & \multicolumn{1}{c|}{\texttt{FT}} & \texttt{Ours} \\ \hline\hline

\parbox[t]{2mm}{\multirow{4}{*}{\rotatebox[origin=c]{90}{\textsc{hopper}}}}
 & \textsc{{{random}}}  & 0.6 $\pm$ 0.6  & 3.8 $\pm$ 1.1  & 3.8 $\pm$ 1.1  & 57.6 $\pm$ 8.7  & \textbf{62.1 $\pm$ 20.9} \\
 & \textsc{{{medium-replay}}}  & 27.0 $\pm$ 4.0  & 72.6 $\pm$ 13.0  & 72.1 $\pm$ 13.6  & 35.2 $\pm$ 4.9  & \textbf{83.6 $\pm$ 8.7} \\
 & \textsc{{{medium}}}  & 30.4 $\pm$ 4.9  & 86.0 $\pm$ 2.9  & \textbf{85.9 $\pm$ 2.9}  & 29.5 $\pm$ 9.6  & 84.7 $\pm$ 5.6 \\
 & \textsc{{{medium-expert}}}  & 12.2 $\pm$ 4.9  & 74.0 $\pm$ 36.4  & 73.9 $\pm$ 36.6  & 33.6 $\pm$ 3.1  & \textbf{77.2 $\pm$ 26.3} \\\hline
 & \textit{Average}  & \textit{17.6 $\pm$ 3.6}  & \textit{59.1 $\pm$ 13.3}  & \textit{58.9 $\pm$ 13.5}  & \textit{39.0 $\pm$ 6.6}  & \textbf{\textit{76.9 $\pm$ 15.4}} \\\hline\hline
\parbox[t]{2mm}{\multirow{4}{*}{\rotatebox[origin=c]{90}{\textsc{cheetah}}}}
 & \textsc{{{random}}}  & 55.8 $\pm$ 2.7  & 91.9 $\pm$ 5.1  & \textbf{91.8 $\pm$ 5.2}  & 76.5 $\pm$ 5.9  & 91.7 $\pm$ 1.3 \\
 & \textsc{{{medium-replay}}}  & 68.4 $\pm$ 7.3  & 84.7 $\pm$ 7.6  & 84.7 $\pm$ 7.6  & 75.4 $\pm$ 5.3  & \textbf{95.8 $\pm$ 2.5} \\
 & \textsc{{{medium}}}  & 46.9 $\pm$ 6.6  & 90.0 $\pm$ 5.9  & 89.8 $\pm$ 6.0  & 56.9 $\pm$ 6.3  & \textbf{92.0 $\pm$ 2.6} \\
 & \textsc{{{medium-expert}}}  & 40.3 $\pm$ 3.2  & 98.7 $\pm$ 0.9  & \textbf{97.8 $\pm$ 1.6}  & 51.0 $\pm$ 6.9  & 97.4 $\pm$ 0.8 \\\hline
 & \textit{Average}  & \textit{52.9 $\pm$ 4.9}  & \textit{91.3 $\pm$ 4.9}  & \textit{91.0 $\pm$ 5.1}  & \textit{65.0 $\pm$ 6.1}  & \textbf{\textit{94.2 $\pm$ 1.8}} \\\hline\hline
\parbox[t]{2mm}{\multirow{4}{*}{\rotatebox[origin=c]{90}{\textsc{walker}}}}
 & \textsc{{{random}}}  & 13.6 $\pm$ 2.6  & 45.1 $\pm$ 4.5  & 41.2 $\pm$ 10.3  & 30.4 $\pm$ 4.6  & \textbf{47.1 $\pm$ 17.2} \\
 & \textsc{{{medium-replay}}}  & 34.1 $\pm$ 8.1  & 95.8 $\pm$ 3.1  & 95.3 $\pm$ 3.8  & 28.7 $\pm$ 6.8  & \textbf{96.3 $\pm$ 2.0} \\
 & \textsc{{{medium}}}  & 24.9 $\pm$ 2.3  & 92.9 $\pm$ 2.9  & \textbf{91.9 $\pm$ 3.7}  & 20.3 $\pm$ 2.1  & 89.9 $\pm$ 2.8 \\
 & \textsc{{{medium-expert}}}  & 22.7 $\pm$ 3.9  & 100.0 $\pm$ 0.0  & \textbf{100.0 $\pm$ 0.0}  & 6.1 $\pm$ 2.1  & 97.5 $\pm$ 2.3 \\\hline
 & \textit{Average}  & \textit{23.8 $\pm$ 4.2}  & \textit{83.4 $\pm$ 2.6}  & \textit{82.1 $\pm$ 4.4}  & \textit{21.4 $\pm$ 3.9}  & \textbf{\textit{82.7 $\pm$ 6.1}} \\\hline\hline
\parbox[t]{2mm}{\multirow{4}{*}{\rotatebox[origin=c]{90}{\textsc{ant}}}}
 & \textsc{{{random}}}  & 9.0 $\pm$ 6.8  & 82.1 $\pm$ 10.4  & 81.4 $\pm$ 10.8  & 5.5 $\pm$ 9.0  & \textbf{83.4 $\pm$ 6.7} \\
 & \textsc{{{medium-replay}}}  & 29.1 $\pm$ 8.9  & 77.4 $\pm$ 13.1  & 77.4 $\pm$ 13.1  & 37.6 $\pm$ 24.1  & \textbf{81.8 $\pm$ 11.2} \\
 & \textsc{{{medium}}}  & 33.6 $\pm$ 5.3  & 69.7 $\pm$ 3.4  & 69.7 $\pm$ 3.4  & 22.8 $\pm$ 1.5  & \textbf{82.3 $\pm$ 3.4} \\
 & \textsc{{{medium-expert}}}  & 47.8 $\pm$ 9.9  & 99.1 $\pm$ 1.5  & \textbf{98.7 $\pm$ 1.4}  & 19.4 $\pm$ 1.9  & 96.6 $\pm$ 2.7 \\\hline
 & Average  & \textit{29.9 $\pm$ 7.7}  & \textit{82.1 $\pm$ 7.1}  & \textit{81.8 $\pm$ 7.2}  & \textit{21.3 $\pm$ 9.1}  & \textbf{\textit{86.0 $\pm$ 6.0}} \\\hline\hline
 & \textit{Overall Average}  & \textit{31.0 $\pm$ 5.1}  & \textit{79.0 $\pm$ 7.0}  & \textit{78.5 $\pm$ 7.6}  & \textit{36.7 $\pm$ 6.4}  & \textbf{\textit{85.0 $\pm$ 7.3}} 
 
\end{tabular}
\end{adjustbox}

\end{center}
\vspace{-1em}
\end{table}

Across all environments and datasets, the \texttt{OPE} scores are typically lower and thus entail higher opportunity loss (regret) relative to the maximum achievable score (i.e., 100\%) with fine-tuning, which underscores the importance of leveraging the online interaction budget. The \texttt{OE} scores are higher mainly because the budget is large enough to obtain reliable and unbiased value estimates for each pretrained policy through online evaluation. However, \texttt{OE} could struggle when the number of candidate policies is large and the budget is insufficient; in such cases, adaptive evaluation approaches (e.g., \cite{konyushova2021active}) can be employed. Nevertheless, the maximum scores attainable by any policy selection approach without fine-tuning--online or offline--is upper-bounded by the values of the best pretrained policies, shown in the \texttt{Best} column. By contrast, as \texttt{FT} scores indicate, straightforward fine-tuning does not substantially improve upon offline algorithms, since some policies may stagnate or regress during fine-tuning, dragging down the average score. 

Our approach combines online evaluation and fine-tuning through an adaptive mechanism, effectively taking the best of both worlds. For example, datasets without good transitions such as \textsc{hopper-random} can yield weak pretrained policies resulting in very low scores, whereas fine-tuning can severely deteriorate high-performing pretrained policies as in \textsc{walker-medium-expert}. Our method monitors whether selected policies are improving with fine-tuning and allocates budget accordingly. If a selected policy is improving, the budget is used for further fine-tuning, avoiding the mistake of spending the entire budget on evaluation, as \texttt{OE} does for \textsc{hopper-random}. If a selected policy is not improving, other candidate policies are selected, resulting in the evaluation of more policies. In this way, the budget is devoted to identifying a high-performing pretrained policy rather than to degrading performance with unnecessary fine-tuning. As a result, by avoiding these pitfalls, our approach achieves the best average scores for each environment and overall.

Finally, increasing the online interaction budget (e.g., from 160 K to 320 K; see Table \ref{table:320}) predictably increases the regret of purely offline approaches, since additional fine-tuning raises the achievable maximum score. However, the differences in scores after doubling the budget are modest, suggesting that offline RL with small interaction budgets is often sufficient to obtain high-performing policies, provided the budget is allocated adaptively and effectively as in our approach. 

\begin{table}[htbp]
\begin{center}

\caption{Results for O2O-RL baselines and our approach under a 320 K online interaction budget.} \label{table:320} 
\begin{adjustbox}{width=0.81\textwidth}
\begin{tabular}{cc|cc|ccc}
\multicolumn{2}{c|}{\multirow{2}{*}{Environment}}  & \multicolumn{2}{c|}{Offline} & \multicolumn{3}{c}{Online} \\ \cline{3-7} 
 & & \multicolumn{1}{c|}{\texttt{OPE}} & \multicolumn{1}{c|}{\texttt{Best}} & \multicolumn{1}{c|}{\texttt{OE}} & \multicolumn{1}{c|}{\texttt{FT}} & \texttt{Ours} \\ \hline

\parbox[t]{2mm}{\multirow{4}{*}{\rotatebox[origin=c]{90}{\textsc{hopper}}}}
 & \textsc{{{random}}}  & 0.5 $\pm$ 0.5  & 2.8 $\pm$ 1.5  & 2.8 $\pm$ 1.5  & 61.5 $\pm$ 9.9  & \textbf{63.3 $\pm$ 13.2} \\
 & \textsc{{{medium-replay}}}  & 24.8 $\pm$ 3.0  & 66.6 $\pm$ 10.0  & 66.1 $\pm$ 10.3  & 61.4 $\pm$ 6.0  & \textbf{80.3 $\pm$ 5.6} \\
 & \textsc{{{medium}}}  & 28.8 $\pm$ 6.2  & 80.8 $\pm$ 7.6  & 80.8 $\pm$ 7.6  & 44.6 $\pm$ 2.3  & \textbf{84.3 $\pm$ 8.4} \\
 & \textsc{{{medium-expert}}}  & 11.6 $\pm$ 4.6  & 70.7 $\pm$ 35.1  & 70.5 $\pm$ 35.3  & 52.7 $\pm$ 2.5  & \textbf{75.4 $\pm$ 22.5} \\\hline
 & \textit{Average}  & \textit{16.4 $\pm$ 3.6}  & \textit{55.2 $\pm$ 13.5}  & \textit{55.0 $\pm$ 13.7}  & \textit{55.0 $\pm$ 5.2}  & \textbf{\textit{75.8 $\pm$ 12.4}} \\\hline\hline
\parbox[t]{2mm}{\multirow{4}{*}{\rotatebox[origin=c]{90}{\textsc{cheetah}}}}
 & \textsc{{{random}}}  & 46.0 $\pm$ 8.5  & 75.1 $\pm$ 9.6  & 75.1 $\pm$ 9.6  & 76.4 $\pm$ 9.4  & \textbf{77.1 $\pm$ 10.8} \\
 & \textsc{{{medium-replay}}}  & 64.1 $\pm$ 9.6  & 79.3 $\pm$ 10.1  & 79.3 $\pm$ 10.1  & 75.8 $\pm$ 7.8  & \textbf{90.7 $\pm$ 6.8} \\
 & \textsc{{{medium}}}  & 46.7 $\pm$ 6.7  & 89.7 $\pm$ 5.5  & 89.6 $\pm$ 5.6  & 72.8 $\pm$ 5.0  & \textbf{91.8 $\pm$ 2.3} \\
 & \textsc{{{medium-expert}}}  & 39.7 $\pm$ 3.4  & 97.1 $\pm$ 3.7  & \textbf{97.1 $\pm$ 3.7}  & 60.4 $\pm$ 4.9  & 95.5 $\pm$ 3.5 \\\hline
 & \textit{Average}  & \textit{49.1 $\pm$ 7.1}  & \textit{85.3 $\pm$ 7.2}  & \textit{85.3 $\pm$ 7.2}  & \textit{71.4 $\pm$ 6.8}  & \textbf{\textit{88.8 $\pm$ 5.8}} \\\hline\hline
\parbox[t]{2mm}{\multirow{4}{*}{\rotatebox[origin=c]{90}{\textsc{walker}}}}
 & \textsc{{{random}}}  & 10.4 $\pm$ 0.5  & 35.1 $\pm$ 3.8  & 35.1 $\pm$ 3.8  & \textbf{50.4 $\pm$ 2.7}  & 40.7 $\pm$ 4.6 \\
 & \textsc{{{medium-replay}}}  & 33.8 $\pm$ 8.3  & 94.9 $\pm$ 3.9  & \textbf{94.9 $\pm$ 3.9}  & 56.4 $\pm$ 2.2  & 92.1 $\pm$ 6.8 \\
 & \textsc{{{medium}}}  & 24.4 $\pm$ 2.5  & 90.9 $\pm$ 3.9  & \textbf{90.0 $\pm$ 4.9}  & 31.4 $\pm$ 2.1  & 89.6 $\pm$ 2.8 \\
 & \textsc{{{medium-expert}}}  & 22.7 $\pm$ 3.9  & 100.0 $\pm$ 0.0  & \textbf{100.0 $\pm$ 0.0}  & 28.0 $\pm$ 3.5  & 96.7 $\pm$ 3.2 \\\hline
 & \textit{Average}  & \textit{22.8 $\pm$ 3.8}  & \textit{80.2 $\pm$ 2.9}  & \textbf{\textit{80.0 $\pm$ 3.2}}  & \textit{41.5 $\pm$ 2.6}  & \textit{79.7 $\pm$ 4.3} \\\hline\hline
\parbox[t]{2mm}{\multirow{4}{*}{\rotatebox[origin=c]{90}{\textsc{ant}}}}
 & \textsc{{{random}}}  & 8.7 $\pm$ 6.5  & 80.1 $\pm$ 9.9  & 79.4 $\pm$ 10.3  & 46.9 $\pm$ 3.9  & \textbf{85.1 $\pm$ 4.6} \\
 & \textsc{{{medium-replay}}}  & 26.6 $\pm$ 6.1  & 73.3 $\pm$ 17.2  & 73.3 $\pm$ 17.2  & 54.7 $\pm$ 5.6  & \textbf{78.9 $\pm$ 11.4} \\
 & \textsc{{{medium}}}  & 33.0 $\pm$ 4.8  & 68.7 $\pm$ 2.0  & 68.7 $\pm$ 2.0  & 48.6 $\pm$ 7.1  & \textbf{82.5 $\pm$ 2.1} \\
 & \textsc{{{medium-expert}}}  & 47.8 $\pm$ 9.9  & 99.1 $\pm$ 1.5  & \textbf{99.1 $\pm$ 1.5}  & 58.2 $\pm$ 9.9  & 97.2 $\pm$ 2.9 \\\hline
 & \textit{Average}  & \textit{29.1 $\pm$ 6.8}  & \textit{80.3 $\pm$ 7.7}  & \textit{80.1 $\pm$ 7.7}  & \textit{52.1 $\pm$ 6.6}  & \textbf{\textit{85.9 $\pm$ 5.2}} \\\hline\hline
 & \textit{Overall Average}  & \textit{29.4 $\pm$ 5.3}  & \textit{75.3 $\pm$ 7.8}  & \textit{75.1 $\pm$ 7.9}  & \textit{55.0 $\pm$ 5.3}  & \textbf{\textit{82.6 $\pm$ 7.0}} 

\end{tabular}
\end{adjustbox}

\end{center}
\vspace{-1em}
\end{table}

\section{Conclusion}

We propose, to our knowledge, the first adaptive O2O-RL framework that jointly performs policy selection and fine-tuning under an online interaction budget. We begin by training a diverse set of candidate policies with offline RL across algorithms and hyperparameters, then use OPE to obtain initial rankings. Acknowledging that pretrained policies can perform arbitrarily poorly and that fine-tuning may stall or regress, depending on the algorithm, hyperparameters, environment, or even random seeds, we allocate scarce online interactions adaptively via a UCB criterion on predicted future values. After each fine-tuning iteration, we evaluate the selected policy, update a linear utoregressive model to refine the value predictions and confidence bounds, and switch when the UCB of another policy exceeds the current one, thereby using the budget efficiently. Across multiple benchmarks, our adaptive selection and fine-tuning procedure consistently improves over O2O-RL baselines.

Although it outperforms strong baselines, our approach leaves ample room for improvement. One limitation is that, after each fine-tuning iteration, it requires a considerable amount of online interaction to evaluate the fine-tuned policy. A future direction could be to leverage exploration rollouts to monitor progress during fine-tuning with little or no explicit online evaluation. Another direction is to incorporate policy similarities, as suggested by \cite{konyushova2021active} to improve efficiency. An orthogonal direction is to develop OPE methods tailored to O2O-RL that focus more on ranking and on the fine-tuning potential of pretrained policies. A further direction is to extend the problem setting to align with the concept of deployment complexity \citep{huang2022towards, su2022muro}.
Lastly, beyond benchmark evaluation, ablation studies, theoretical analysis, and connections to other paradigms such as Hybrid RL \citep{song2023hybrid} could provide further understanding and guide future improvement.
We believe our framework will contribute to offline RL research by taking one more step toward practical, deployable recipes for real-world systems where online interactions are costly or risky.

\acks{
This work is sponsored in part by the NSF under NAIAD Award 2332744, the National AI Institute for Edge Computing Leveraging Next Generation Wireless Networks under Grant CNS-2112562, as well as the Commonwealth Cyber Initiative's Central Virginia Node under the awards VV-1Q26-001, HV-2025-035, and HC-2025-033.
We are deeply grateful to all the reviewers for their thoughtful and constructive feedback.
}

\bibliography{l4dc2026}

\begin{thebibliography}{48}
\providecommand{\natexlab}[1]{#1}
\providecommand{\url}[1]{\texttt{#1}}
\expandafter\ifx\csname urlstyle\endcsname\relax
  \providecommand{\doi}[1]{doi: #1}\else
  \providecommand{\doi}{doi: \begingroup \urlstyle{rm}\Url}\fi

\bibitem[Ball et~al.(2023)Ball, Smith, Kostrikov, and Levine]{ball2023efficient}
Philip~J Ball, Laura Smith, Ilya Kostrikov, and Sergey Levine.
\newblock Efficient online reinforcement learning with offline data.
\newblock In \emph{International Conference on Machine Learning}, pages 1577--1594. PMLR, 2023.

\bibitem[Bollerslev(1986)]{bollerslev1986generalized}
Tim Bollerslev.
\newblock Generalized autoregressive conditional heteroskedasticity.
\newblock \emph{Journal of econometrics}, 31\penalty0 (3):\penalty0 307--327, 1986.

\bibitem[Box et~al.(2015)Box, Jenkins, Reinsel, and Ljung]{box2015time}
George~EP Box, Gwilym~M Jenkins, Gregory~C Reinsel, and Greta~M Ljung.
\newblock \emph{Time series analysis: forecasting and control}.
\newblock John Wiley \& Sons, 2015.

\bibitem[Brandfonbrener et~al.(2021)Brandfonbrener, Whitney, Ranganath, and Bruna]{brandfonbrener2021offline}
David Brandfonbrener, Will Whitney, Rajesh Ranganath, and Joan Bruna.
\newblock Offline rl without off-policy evaluation.
\newblock \emph{Advances in neural information processing systems}, 34:\penalty0 4933--4946, 2021.

\bibitem[Campanaro et~al.(2024)Campanaro, Gangapurwala, Merkt, and Havoutis]{campanaro2024learning}
Luigi Campanaro, Siddhant Gangapurwala, Wolfgang Merkt, and Ioannis Havoutis.
\newblock Learning and deploying robust locomotion policies with minimal dynamics randomization.
\newblock In \emph{6th Annual Learning for Dynamics \& Control Conference}, pages 578--590. PMLR, 2024.

\bibitem[Chang et~al.(2025)Chang, Ballentine, He, Kim, Jiang, Liang, Palacios, Wang, Piacenza, Kymissis, et~al.]{chang2025spikeatac}
Eric~T Chang, Peter Ballentine, Zhanpeng He, Do-Gon Kim, Kai Jiang, Hua-Hsuan Liang, Joaquin Palacios, William Wang, Pedro Piacenza, Ioannis Kymissis, et~al.
\newblock Spikeatac: A multimodal tactile finger with taxelized dynamic sensing for dexterous manipulation.
\newblock \emph{arXiv preprint arXiv:2510.27048}, 2025.

\bibitem[Coumans and Bai(2021)]{coumans2021}
Erwin Coumans and Yunfei Bai.
\newblock Pybullet, a python module for physics simulation for games, robotics and machine learning.
\newblock \url{http://pybullet.org}, 2021.
\newblock Accessed 10 November 2025.

\bibitem[Dulac-Arnold et~al.(2021)Dulac-Arnold, Levine, Mankowitz, Li, Paduraru, Gowal, and Hester]{dulac2021challenges}
Gabriel Dulac-Arnold, Nir Levine, Daniel~J Mankowitz, Jerry Li, Cosmin Paduraru, Sven Gowal, and Todd Hester.
\newblock Challenges of real-world reinforcement learning: definitions, benchmarks and analysis.
\newblock \emph{Machine Learning}, 110\penalty0 (9):\penalty0 2419--2468, 2021.

\bibitem[Engle(1982)]{engle1982autoregressive}
Robert~F Engle.
\newblock Autoregressive conditional heteroscedasticity with estimates of the variance of united kingdom inflation.
\newblock \emph{Econometrica: Journal of the econometric society}, pages 987--1007, 1982.

\bibitem[Farama(2021)]{farama2021}
Foundation Farama.
\newblock Dataset reproducibility guide.
\newblock \url{https://github.com/Farama-Foundation/d4rl/wiki/Dataset-Reproducibility-Guide}, 2021.
\newblock Accessed 10 November 2025.

\bibitem[Feng et~al.(2023)Feng, Sun, Yan, Zhu, Zou, Shen, and Liu]{feng2023dense}
Shuo Feng, Haowei Sun, Xintao Yan, Haojie Zhu, Zhengxia Zou, Shengyin Shen, and Henry~X Liu.
\newblock Dense reinforcement learning for safety validation of autonomous vehicles.
\newblock \emph{Nature}, 615\penalty0 (7953):\penalty0 620--627, 2023.

\bibitem[Florence et~al.(2022)Florence, Lynch, Zeng, Ramirez, Wahid, Downs, Wong, Lee, Mordatch, and Tompson]{florence2022implicit}
Pete Florence, Corey Lynch, Andy Zeng, Oscar~A Ramirez, Ayzaan Wahid, Laura Downs, Adrian Wong, Johnny Lee, Igor Mordatch, and Jonathan Tompson.
\newblock Implicit behavioral cloning.
\newblock In \emph{Conference on robot learning}, pages 158--168. PMLR, 2022.

\bibitem[Francq and Zakoian(2019)]{francq2019garch}
Christian Francq and Jean-Michel Zakoian.
\newblock \emph{GARCH models: structure, statistical inference and financial applications}.
\newblock John Wiley \& Sons, 2019.

\bibitem[Fu et~al.(2020)Fu, Kumar, Nachum, Tucker, and Levine]{fu2020d4rl}
Justin Fu, Aviral Kumar, Ofir Nachum, George Tucker, and Sergey Levine.
\newblock D4rl: Datasets for deep data-driven reinforcement learning.
\newblock \emph{arXiv preprint arXiv:2004.07219}, 2020.

\bibitem[Fujimoto and Gu(2021)]{fujimoto2021minimalist}
Scott Fujimoto and Shixiang~Shane Gu.
\newblock A minimalist approach to offline reinforcement learning.
\newblock \emph{Advances in neural information processing systems}, 34:\penalty0 20132--20145, 2021.

\bibitem[Fujimoto et~al.(2018)Fujimoto, Hoof, and Meger]{fujimoto2018addressing}
Scott Fujimoto, Herke Hoof, and David Meger.
\newblock Addressing function approximation error in actor-critic methods.
\newblock In \emph{International conference on machine learning}, pages 1587--1596. PMLR, 2018.

\bibitem[Fujimoto et~al.(2019)Fujimoto, Meger, and Precup]{fujimoto2019off}
Scott Fujimoto, David Meger, and Doina Precup.
\newblock Off-policy deep reinforcement learning without exploration.
\newblock In \emph{International conference on machine learning}, pages 2052--2062. PMLR, 2019.

\bibitem[Haarnoja et~al.(2018)Haarnoja, Zhou, Abbeel, and Levine]{haarnoja2018soft}
Tuomas Haarnoja, Aurick Zhou, Pieter Abbeel, and Sergey Levine.
\newblock Soft actor-critic: Off-policy maximum entropy deep reinforcement learning with a stochastic actor.
\newblock In \emph{International Conference on Machine Learning}, pages 1861--1870. PMLR, 2018.

\bibitem[Huang et~al.(2022)Huang, Chen, Zhao, Qin, Jiang, and Liu]{huang2022towards}
Jiawei Huang, Jinglin Chen, Li~Zhao, Tao Qin, Nan Jiang, and Tie-Yan Liu.
\newblock Towards deployment-efficient reinforcement learning: Lower bound and optimality.
\newblock In \emph{International Conference on Learning Representations}, 2022.
\newblock URL \url{https://openreview.net/forum?id=ccWaPGl9Hq}.

\bibitem[Julian et~al.(2021)Julian, Swanson, Sukhatme, Levine, Finn, and Hausman]{julian2021never}
Ryan Julian, Benjamin Swanson, Gaurav Sukhatme, Sergey Levine, Chelsea Finn, and Karol Hausman.
\newblock Never stop learning: The effectiveness of fine-tuning in robotic reinforcement learning.
\newblock In \emph{Conference on Robot Learning}, pages 2120--2136. PMLR, 2021.

\bibitem[Konyushova et~al.(2021)Konyushova, Chen, Paine, Gulcehre, Paduraru, Mankowitz, Denil, and de~Freitas]{konyushova2021active}
Ksenia Konyushova, Yutian Chen, Thomas Paine, Caglar Gulcehre, Cosmin Paduraru, Daniel~J Mankowitz, Misha Denil, and Nando de~Freitas.
\newblock Active offline policy selection.
\newblock \emph{Advances in Neural Information Processing Systems}, 34:\penalty0 24631--24644, 2021.

\bibitem[Kostrikov et~al.(2021)Kostrikov, Nair, and Levine]{kostrikov2021offline}
Ilya Kostrikov, Ashvin Nair, and Sergey Levine.
\newblock Offline reinforcement learning with implicit q-learning.
\newblock In \emph{Deep RL Workshop NeurIPS}, 2021.

\bibitem[Kumar et~al.(2019)Kumar, Fu, Soh, Tucker, and Levine]{kumar2019stabilizing}
Aviral Kumar, Justin Fu, Matthew Soh, George Tucker, and Sergey Levine.
\newblock Stabilizing off-policy q-learning via bootstrapping error reduction.
\newblock \emph{Advances in neural information processing systems}, 32, 2019.

\bibitem[Kumar et~al.(2020)Kumar, Zhou, Tucker, and Levine]{kumar2020conservative}
Aviral Kumar, Aurick Zhou, George Tucker, and Sergey Levine.
\newblock Conservative q-learning for offline reinforcement learning.
\newblock \emph{Advances in neural information processing systems}, 33:\penalty0 1179--1191, 2020.

\bibitem[Kurenkov and Kolesnikov(2022)]{kurenkov2022showing}
Vladislav Kurenkov and Sergey Kolesnikov.
\newblock Showing your offline reinforcement learning work: Online evaluation budget matters.
\newblock In \emph{International Conference on Machine Learning}, pages 11729--11752. PMLR, 2022.

\bibitem[Ladosz et~al.(2022)Ladosz, Weng, Kim, and Oh]{ladosz2022exploration}
Pawel Ladosz, Lilian Weng, Minwoo Kim, and Hyondong Oh.
\newblock Exploration in deep reinforcement learning: A survey.
\newblock \emph{Information Fusion}, 85:\penalty0 1--22, 2022.

\bibitem[Le et~al.(2019)Le, Voloshin, and Yue]{le2019batch}
Hoang Le, Cameron Voloshin, and Yisong Yue.
\newblock Batch policy learning under constraints.
\newblock In \emph{International Conference on Machine Learning}, pages 3703--3712. PMLR, 2019.

\bibitem[Lee et~al.(2022)Lee, Seo, Lee, Abbeel, and Shin]{lee2022offline}
Seunghyun Lee, Younggyo Seo, Kimin Lee, Pieter Abbeel, and Jinwoo Shin.
\newblock Offline-to-online reinforcement learning via balanced replay and pessimistic q-ensemble.
\newblock In \emph{Conference on Robot Learning}, pages 1702--1712. PMLR, 2022.

\bibitem[Levine et~al.(2020)Levine, Kumar, Tucker, and Fu]{levine2020offline}
Sergey Levine, Aviral Kumar, George Tucker, and Justin Fu.
\newblock Offline reinforcement learning: Tutorial, review, and perspectives on open problems.
\newblock \emph{arXiv preprint arXiv:2005.01643}, 2020.

\bibitem[Lillicrap et~al.(2016)Lillicrap, Hunt, Pritzel, Heess, Erez, Tassa, Silver, and Wierstra]{lillicrap2016continuous}
Timothy~P Lillicrap, Jonathan~J Hunt, Alexander Pritzel, Nicolas Heess, Tom Erez, Yuval Tassa, David Silver, and Daan Wierstra.
\newblock Continuous control with deep reinforcement learning.
\newblock In \emph{International Conference on Learning Representations}, 2016.

\bibitem[Nair et~al.(2020)Nair, Gupta, Dalal, and Levine]{nair2020awac}
Ashvin Nair, Abhishek Gupta, Murtaza Dalal, and Sergey Levine.
\newblock Awac: Accelerating online reinforcement learning with offline datasets.
\newblock \emph{arXiv preprint arXiv:2006.09359}, 2020.

\bibitem[Nakamoto et~al.(2023)Nakamoto, Zhai, Singh, Sobol~Mark, Ma, Finn, Kumar, and Levine]{nakamoto2023cal}
Mitsuhiko Nakamoto, Simon Zhai, Anikait Singh, Max Sobol~Mark, Yi~Ma, Chelsea Finn, Aviral Kumar, and Sergey Levine.
\newblock Cal-ql: Calibrated offline rl pre-training for efficient online fine-tuning.
\newblock \emph{Advances in Neural Information Processing Systems}, 36:\penalty0 62244--62269, 2023.

\bibitem[Paine et~al.(2020)Paine, Paduraru, Michi, Gulcehre, Zolna, Novikov, Wang, and de~Freitas]{paine2020hyperparameter}
Tom~Le Paine, Cosmin Paduraru, Andrea Michi, Caglar Gulcehre, Konrad Zolna, Alexander Novikov, Ziyu Wang, and Nando de~Freitas.
\newblock Hyperparameter selection for offline reinforcement learning.
\newblock \emph{arXiv preprint arXiv:2007.09055}, 2020.

\bibitem[Prudencio et~al.(2023)Prudencio, Maximo, and Colombini]{prudencio2023survey}
Rafael~Figueiredo Prudencio, Marcos~ROA Maximo, and Esther~Luna Colombini.
\newblock A survey on offline reinforcement learning: Taxonomy, review, and open problems.
\newblock \emph{IEEE Transactions on Neural Networks and Learning Systems}, 35\penalty0 (8):\penalty0 10237--10257, 2023.

\bibitem[Qin et~al.(2022)Qin, Zhang, Gao, Chen, Li, Zhang, and Yu]{qin2022neorl}
Rong-Jun Qin, Xingyuan Zhang, Songyi Gao, Xiong-Hui Chen, Zewen Li, Weinan Zhang, and Yang Yu.
\newblock Neorl: A near real-world benchmark for offline reinforcement learning.
\newblock \emph{Advances in Neural Information Processing Systems}, 35:\penalty0 24753--24765, 2022.

\bibitem[Seno and Imai(2022)]{d3rlpy}
Takuma Seno and Michita Imai.
\newblock d3rlpy: An offline deep reinforcement learning library.
\newblock \emph{Journal of Machine Learning Research}, 23\penalty0 (315):\penalty0 1--20, 2022.
\newblock URL \url{http://jmlr.org/papers/v23/22-0017.html}.

\bibitem[Sheppard(2021)]{arch}
Kevin Sheppard.
\newblock Univariate volatility modeling, bootstrapping, multiple comparison procedures and unit root tests.
\newblock \url{https://github.com/bashtage/arch}, 2021.
\newblock Accessed 10 November 2025.

\bibitem[Singh et~al.(2022)Singh, Kumar, and Singh]{singh2022reinforcement}
Bharat Singh, Rajesh Kumar, and Vinay~Pratap Singh.
\newblock Reinforcement learning in robotic applications: a comprehensive survey.
\newblock \emph{Artificial Intelligence Review}, 55\penalty0 (2):\penalty0 945--990, 2022.

\bibitem[Song et~al.(2023)Song, Zhou, Sekhari, Bagnell, Krishnamurthy, and Sun]{song2023hybrid}
Yuda Song, Yifei Zhou, Ayush Sekhari, Drew Bagnell, Akshay Krishnamurthy, and Wen Sun.
\newblock Hybrid {RL}: Using both offline and online data can make {RL} efficient.
\newblock In \emph{The Eleventh International Conference on Learning Representations}, 2023.
\newblock URL \url{https://openreview.net/forum?id=yyBis80iUuU}.

\bibitem[Su et~al.(2022)Su, Lee, Mulvey, and Poor]{su2022muro}
DiJia Su, Jason~D. Lee, John Mulvey, and H.~Vincent Poor.
\newblock {MURO}: Deployment constrained reinforcement learning with model-based uncertainty regularized batch optimization, 2022.
\newblock URL \url{https://openreview.net/forum?id=eWNpRVcfzi}.

\bibitem[Tang et~al.(2025)Tang, Abbatematteo, Hu, Chandra, Mart{\'\i}n-Mart{\'\i}n, and Stone]{tang2025deep}
Chen Tang, Ben Abbatematteo, Jiaheng Hu, Rohan Chandra, Roberto Mart{\'\i}n-Mart{\'\i}n, and Peter Stone.
\newblock Deep reinforcement learning for robotics: A survey of real-world successes.
\newblock In \emph{Proceedings of the AAAI Conference on Artificial Intelligence}, volume~39, pages 28694--28698, 2025.

\bibitem[Tarasov et~al.(2023)Tarasov, Kurenkov, Nikulin, and Kolesnikov]{tarasov2023revisiting}
Denis Tarasov, Vladislav Kurenkov, Alexander Nikulin, and Sergey Kolesnikov.
\newblock Revisiting the minimalist approach to offline reinforcement learning.
\newblock \emph{Advances in Neural Information Processing Systems}, 36:\penalty0 11592--11620, 2023.

\bibitem[Tian et~al.(2024)Tian, Zhao, Lin, Flynn, Zhao, and Tian]{tian2024balanced}
Zhen Tian, Dezong Zhao, Zhihao Lin, David Flynn, Wenjing Zhao, and Daxin Tian.
\newblock Balanced reward-inspired reinforcement learning for autonomous vehicle racing.
\newblock In \emph{6th Annual Learning for Dynamics \& Control Conference}, pages 628--640. PMLR, 2024.

\bibitem[Torabi et~al.(2018)Torabi, Warnell, and Stone]{torabi2018behavioral}
Faraz Torabi, Garrett Warnell, and Peter Stone.
\newblock Behavioral cloning from observation.
\newblock In \emph{International Joint Conference on Artificial Intelligence}, pages 4950--4957, 2018.

\bibitem[Uehara et~al.(2025)Uehara, Shi, and Kallus]{uehara2025review}
Masatoshi Uehara, Chengchun Shi, and Nathan Kallus.
\newblock A review of off-policy evaluation in reinforcement learning.
\newblock \emph{Statistical Science}, 2025.

\bibitem[Zhang et~al.(2022)Zhang, Lv, Li, Bao, Liu, and Liu]{zhang2022reinforcement}
Rong Zhang, Qibing Lv, Jie Li, Jinsong Bao, Tianyuan Liu, and Shimin Liu.
\newblock A reinforcement learning method for human-robot collaboration in assembly tasks.
\newblock \emph{Robotics and Computer-Integrated Manufacturing}, 73:\penalty0 102227, 2022.

\bibitem[Zhou et~al.(2023)Zhou, Ke, Srinivasa, Gupta, Rajeswaran, and Kumar]{zhou2023real}
Gaoyue Zhou, Liyiming Ke, Siddhartha Srinivasa, Abhinav Gupta, Aravind Rajeswaran, and Vikash Kumar.
\newblock Real world offline reinforcement learning with realistic data source.
\newblock In \emph{2023 IEEE International Conference on Robotics and Automation (ICRA)}, pages 7176--7183. IEEE, 2023.

\bibitem[Zhou et~al.(2021)Zhou, Bajracharya, and Held]{zhou2021plas}
Wenxuan Zhou, Sujay Bajracharya, and David Held.
\newblock Plas: Latent action space for offline reinforcement learning.
\newblock In \emph{Conference on Robot Learning}, pages 1719--1735. PMLR, 2021.

\end{thebibliography}

\end{document}